# A Mathematical Model, Implementation and Study of a Swarm System

Blesson Varghese and Gerard T. McKee

*Abstract*—The work reported in this paper is motivated towards the development of a mathematical model for swarm systems based on macroscopic primitives. A pattern formation and transformation model is proposed. The pattern transformation model comprises two general methods for pattern transformation, namely a macroscopic transformation and mathematical transformation method. The problem of transformation is formally expressed and four special cases of transformation are considered. Simulations to confirm the feasibility of the proposed models and transformation methods are presented. Comparison between the two transformation methods is also reported.

## I. INTRODUCTION

SWARM robotics research specifically deals with some classical problems which have been of concern to researchers over the years. An overall review of work done in swarm robotics presented in [1] identifies pattern formation as one challenge. The problem of pattern formation relates to the microscopic (influencing individual robot behavior) or macroscopic (influencing group behavior) properties of a swarm system.

Pattern formation has been addressed by researchers in swarm robotics using two approaches, namely potential fields and behavior based models. In [2] a control law is proposed for pattern formation that consists of the sum of a repulsive potential and an attractive potential field. The approach is useful for the classical problem of obstacle avoidance in patterns. The paper reports pattern formation control based on parameters describing microscopic behaviors and presents a mathematical model for the control strategy. In [3, 4] potential field and sliding mode control have been used for pattern formations. In [5] the authors study a general class of Attractive and Repulsive functions used to achieve swarm aggregation. These models lack a description of a swarm model describing the macroscopic properties.

The behavior based approach has focused on pattern formation behaviors for multi-robot teams and patterns including the line, column, diamond and wedge geometric formations have been reported [6]. Obstacle avoidance and other navigational behaviors are integrated within the model. Though the microscopic properties of the system are defined, the group behavior of a system cannot be explicitly determined by the behavior based approach. Moreover this approach does not lend itself to mathematical analysis and formulations. Hence, the approach fails in articulating a swarm model with macroscopic parameters.

Other approaches used by researchers in pattern formation include the dynamic window approach [7] – [9] and flow field method [10]. Both these approaches consider microscopic properties of the swarm system.

In short, pattern formation approaches do not consider macroscopic parameters of a swarm system. Hence there arises a need to develop a pattern formation model based on macroscopic parameters. There are five main benefits of using macroscopic parameters. Firstly, implicit coordination, which refers to the coordination of a pattern comprising of mobile robots, need not be specified externally. Coordination is achieved as a result of varying the macroscopic properties. Secondly, Group behavior definition, which refers to the collective behavior of the group, is possible by controlling the macroscopic parameters. The individual behavior of the units is affected by the variation in the macroscopic property. Thirdly, Adaptability, which refers to the ability of the group to adjust to change of internal or external circumstances, can be affected by macroscopic parameters. Fourthly, Stability, which refers to the factor by which the robot group maintains a pattern, can be controlled by using macroscopic parameter to dampen the propagation of errors. Fifthly, higher order parameters can control parameters of lower order.

Researchers tend to consider pattern transformation along with pattern formation without much distinction. Transformation, which refers to the reconfiguration of swarm patterns, is a little considered area within swarm robotics. Transformation of patterns is an appropriate response to obstacles for unhindered motion. Patterns are reconfigured by repositioning all or a subset of agents in a swarm. A transformation may result in a change of geometric orientation of a pattern and relationships between interacting units in the pattern.

Work based on pattern transformation is reported in [11], where a stable virtual leader pattern transforms to a different pattern by the addition of a morphing force. Illustrations of transformation and mathematical notations for computation of forces in the pattern are also presented. The transformation technique facilitates pattern change by allowing participating agents to find their own equilibrium. However, the morphing procedure for transforming pattern is not defined.

An algorithm reported in [12] is capable of transforming patterns in response to a command issued by a human operator. The command is issued to a single robot and causes a chain reaction in the neighboring robots resulting in a global transformation. Pattern transformation from a parabola to a sine curve is illustrated. Though the notion of transforming patterns is presented, the transformation method remains unaddressed.

A relative distance versus orientation model for transformation is reported in [13]. The strategy involves varying the orientation value to globally transform a triangle to a line formation. Though a positional transformation is not executed on all participating agents (one position remains unaltered), a global geometrical transformation is achieved. This strategy is specific to the scenario when a triangle to line transformation is performed.

The use of affine transformations, a mathematical tool, for transforming swarms patterns is presented in [14]. The target position of each member of the swarm is pre-determined by the mathematical transformation tool. The shortest path between the original and target positions is traced by considering the ant colony optimization [15] algorithm. Simulation results illustrate the transformation of a horizontal line pattern to a diagonal line pattern. Though a mathematical method is explored, a geometrical transformation between different shapes may not be possible using affine transformations.

In short, researchers have not concentrated on investigating pattern transformation in swarm robotics. General methods for reconfiguring patterns are not emphasized either.

The work reported in this paper is motivated towards the development of a swarm model based on primary and secondary macroscopic primitives. A pattern transformation model comprising two transformation methods that enable geometric transformation are proposed. Firstly, a swarm macroscopic parameter oriented method based on macroscopic (group behavior of a swarm pattern) parameters is proposed. Secondly, a mathematical tool based on macroscopic and microscopic (individual robot behavior) parameters is proposed. The problem of transformation is formally stated and four special cases of transformation are considered. Experimental studies confirming the feasibility of the proposed methods are presented.

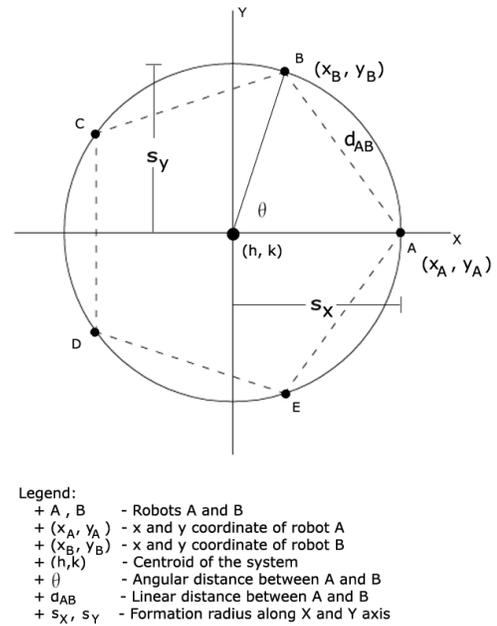

Legend:
+ A , B       - Robots A and B
+ ($x_A$, $y_A$) - x and y coordinate of robot A
+ ($x_B$, $y_B$) - x and y coordinate of robot B
+ (h,k)       - Centroid of the system
+ $\theta$    - Angular distance between A and B
+ $d_{AB}$    - Linear distance between A and B
+ $s_X$, $s_Y$ - Formation radius along X and Y axis

Fig. 1. The primitives of the multi-agent pattern formation model.

The remainder of this paper is organized as follows. Section II proposes the pattern formation model. Section III states the problem of transformation and presents the pattern transformation model. Sections IV and V proposes two transformation methods comprising the pattern transformation model. The feasibility of the proposed models is confirmed by simulation which is reported in Section VI. Section VII concludes this paper and reports future work.

II. SWARM PATTERN FORMATION MODEL

This section proposes a mathematical model for swarm pattern formation based on primary and secondary primitives. The mathematical model is formulated based on the foundations of the Complex Plane. The De Moivre's formula to obtain roots of an equation is used to represent the model. If z = x + iy [16] and is represented in the polar form as $z = r(\cos\theta + i\sin\theta)$ and r is called the absolute value or the modulus of z, then $z^n = r^n(\cos n\theta + i\sin n\theta)$ for n = 0, 1, 2,… The $n^{th}$ root of z is obtained by $\sqrt[n]{z} = \sqrt[n]{r}\left(\cos\left[\dfrac{\theta + 2k\pi}{n}\right] + i\sin\left[\dfrac{\theta + 2k\pi}{n}\right]\right)$ where k = 0, 1,… (n-1). The roots of the equation lie on a circle of radius

$\sqrt[n]{r}$ with centre at the origin and constitute the vertices of a regular polygon of n sides. The result of joining the n roots is an n-sided polygon. The polygon is circumscribed by a circle otherwise referred as the circumcircle of the polygon. When mapping these results onto the area of multi-agent pattern formations, it is assumed that the robotic agents are positioned on the vertices of the polygon. Hence the robots form a closed polygonal pattern and the system is mobile with appropriate communication and coordination mechanism.

The mathematical model is realized by considering macroscopic primitives (Figure 1). The term primitive in this paper refers to an element used as a building block to define aspects of the model. The macroscopic primitives are separated into primary and secondary primitives. Primary macroscopic primitives are basic or fundamental elements. They are considered as input variables to the model and are irreducible to simpler parameters or expressions and therefore termed as independent primitives. Secondary macroscopic primitives are derived from other primitives of the mathematical model. Hence, these primitives are termed as dependent primitives.

The primary macroscopic primitives of the model proposed in this paper are the total number of robots, angular separation, formation radius and elongation. The total number of robots in a polygonal pattern, given by n, equates to the number of vertices of a polygon or the roots of the complex equation. Angular separation is an important factor that determines the coordinates for positioning robots in a polygonal pattern. Angular separation, denoted by $\theta$, is a measure of the angular spacing between adjacent robots of a pattern. Formation radius, denoted by r, is the radius of the circumscribing circle of the polygonal pattern. This primitive determines the area occupied by the pattern. Elongation ratio of a pattern, denoted by e, is a ratio of magnitudes of the major and minor axis of the pattern and quantifies the shape transforming behavior of a pattern. The symmetry of a pattern can also be described by the elongation ratio.

The secondary macroscopic primitives are linear distance and shaping radii. The distance between adjacent robots in the polygon is a constant if the polygon is regular. To compute the distance between robots, the coordinate positions of the robots need to be known. The centroid of the pattern, (h,k), is used to compute the coordinates of robots. Further, the Euclidean distance between adjacent robots A and B is given by $d_{AB} = \sqrt{(x_B - x_A)^2 + (y_B - y_A)^2}$ . Hence, linear distance is dependent on the position coordinates of robots.

The shaping radii along the x and y axis, $s_x$ and $s_y$ respectively determine the measure of deflation or inflation of a pattern laterally and longitudinally. The magnitudes of elongation and formation radius are useful to determine the shaping radii of a pattern and are given by $s_x = re$ and $s_y = \dfrac{r}{e}$. The equations that define the shaping radii are also given by $x_B = h + s_x \cos\theta$ and $y_B = k + s_y \sin\theta$ . Hence, orientation radii are dependent on formation radius and elongation.

Variation in the magnitude of the shaping radii results in a regular or irregular pattern. Regular patterns refer to swarm formations as regular polygons. The regularity of polygonal patterns is preserved by scaling (deflate or inflate) the pattern laterally and longitudinally in equal magnitudes. On the other hand, irregular patterns refer to swarm formations as irregular polygons. Irregular patterns can be obtained by scaling patterns laterally or longitudinally with unequal magnitudes.

The swarm pattern formation model presented here is chosen for the study of transformation. Two pattern transformation methods are applied to the swarm model; these are discussed in Section IV and V.

### III. SWARM PATTERN TRANSFORMATION MODEL

Considering the fact that pattern transformation is little addressed in research and general methods for transformations are not investigated, the problem of swarm pattern transformation is presented here.

The term transformation is also associated with modular robotic systems. Algorithms to transform the shape of modular robots are reported in [17]-[19]. However, it is necessary to draw distinction between reconfiguration in modular systems and transformation of swarm patterns considered in this paper. Firstly, in modular robot systems physical connectivity between modules exists ensuring modules in close vicinity of adjacent modules. Secondly, reconfiguring in modules is constrained by being able to reposition on the periphery of an attached module. Thirdly, reconfiguration in modules is not strictly geometry oriented.

Definition: Consider a pattern $P$ with geometric relationships represented as $G_P$. The pattern $P$ comprises of N robots such that their positions are given by $p_i(x_i, y_i)$ where $p_i \in \Re^2$ and $i = 1, 2, ..., N$. Pattern $P$ transforms into the pattern $Q$ with geometric constraints or relationships represented as $G_Q$. The pattern $Q$ also comprises of N robots such that the position of the robotic agents is given by $q_i(x_i, y_i)$ where $q_i \in \Re^2$ and $i = 1,2,...,N$.

The function which enables the transformation of the pattern $P$ to $Q$ is given by $f(P) = Q$. In other words,
$$f(p_1(x_1, y_1), p_2(x_2, y_2),...,p_N(x_N, y_N)) = q_1(x_1, y_1), q_2(x_2, y_2),...,q_N(x_N, y_N).$$
The application of an inverse transformation function on the transformed pattern $Q$ yields the pattern $P$, given by $f^{-1}(Q) = P$. The transformation on the pattern also results in a transformation of the geometrical relationships from $G_P$ to $G_Q$ between the participating agents in the pattern. Four cases of transformation based on the above definition are derived by imposing restrictions on the geometrical constraints.

*Case 1:* $G_Q = G_P$ after a transformation that involves repositioning all agents. This case is relevant when robotic agents in the pattern have repositioned, yet the geometrical pattern has not changed. Such a transformation is termed as Elementary transformation in this paper. This term also refers to those transformations very basic in nature. For instance, a swarm could be rotated with respect to its centroid or translated such that all robotic agents have repositioned themselves. Though the orientation of the pattern has changed, the configuration of the pattern remains unaltered. Mathematically, the case of elementary transformation would be such that $G_Q = G_P$ and $\forall i : p_i(x_i, y_i) \neq q_i(x_i, y_i)$.

*Case 2:* $G_Q = G_P$ after a transformation without repositioning all agents. This case considers the rotation or translation of the swarm with respect to some robotic agent whose position remains fixed. This case is also classified under Elementary transformation, yet repositioning of all agents has not occurred. Mathematically, this case of elementary transformation would be such that $G_Q = G_P$ and $\exists i : p_i(x_i, y_i) = q_i(x_i, y_i)$

*Case 3:* $G_Q \neq G_P$ after a transformation that involves repositioning all agents. This relates to the case when the geometrical constraints of the pattern have changed and a new pattern has emerged. It is termed a Geometric transformation. This concept is relevant when robotic agents in the pattern reposition to result in a geometry change. For instance, the shape of a swarm could be geometrically transformed from a polygon to a line. It is interesting to note that the scaling of a pattern would result in a geometric transformation, since the geometrical constraints are dissimilar in both cases. Mathematically, the case of geometrical transformation would be such that $G_Q \neq G_P$ and $\forall i : p_i(x_i, y_i) \neq q_i(x_i, y_i)$.

*Case 4:* $G_Q \neq G_P$ after transformation without repositioning all agents. This case considers the geometric transformation such that the position of one or more than one robotic agent remains fixed. It is classified under geometric transformation, yet repositioning of all agents has not occurred. Mathematically, the case of geometrical transformation would be such that $G_Q \neq G_P$ and $\exists i : p_i(x_i, y_i) = q_i(x_i, y_i)$.

Cases 1 and 2 relate to elementary transformation of the pattern. In these cases, the geometric constraint or relationship persists even after elementary transformation. Cases 3 and 4 consider geometric transformation. In these cases the geometrical relationships change after transformation.

The swarm pattern formation model presented in the previous section is chosen for the study of transformation. Two feasible transformation methods, namely a macroscopic transformation method and a mathematical transformation method are proposed in this paper. Cases 1, 3 and 4 of transformation are considered in the transformation methods. Case 2 will be reported in a future paper.

IV. METHOD 1: MACROSCOPIC TRANSFORMATION

The transformation method proposed in this section is inclusive of elementary and geometric transformations applied on the swarm model. Transformations of cases 1, 3 and 4 are achieved by varying the secondary macroscopic primitives, namely the shaping radii (along x and y axis) of the swarm model. It is interesting to note that a sequences of operations performed on the swarm model results in a transformation. The set of operations are:

*1) Rotation*: The initial step of rotation of the model is performed to achieve collision avoidance during the next step. A predefined angle offset is used to rotate the swarm. Though the robots are repositioned, the operation results in the same polygonal pattern with a different orientation from the former. Here, the concept of elementary transformation is introduced. Though all robots were repositioned in this operation, a geometric transformation is not evident since the shape of the pattern is retained. Though a geometric transformation is not evident, yet an elementary transformation of case 1 is achieved in this step.

*2) Macroscopic Parameter Operation*: Following a rotation operation, the macroscopic parameter is set to be modified. Deflating the model along the y-axis would result in a deformed polygonal pattern. The deflation of the model is performed by decrementing the magnitude of the shaping radius along the y-axis. When deflation has reached its maximum value, the robotic agents have aligned themselves entirely along the x-axis. Maximum deflation is achieved when the shaping radius value along any axis vanishes. An inflation operation along the other perpendicular axis simultaneously while deflating would result in a pattern with larger inter-linear distance between the agents (a measure for avoiding collisions). This variation is possible due to the notion of flexibility in rigid patterns.

*3) Further Rotation*: This step is performed to achieve equidistance between the participating agents. Though the pattern has transformed its shape by this step, the participating agents are still governed by the rules of the swarm model. A corrective rotation measure would ensure that the agents are loosely equidistant.

## V. METHOD 2: MATHEMATICAL TRANSFORMATION

The method proposed in this section considers case 3, which is achieved by using a mathematical transformation tool. Many mathematical tools are available for transformations which include stretching, rotating, reflecting and translating transformations. The linear fractional transformation is one such readily available mapping function that maps a set of points from one plane to another. The transformation is given by $f(z) = \frac{az+b}{cz+d}$, where z, a, b, c and d are complex numbers satisfying $ad - bc \neq 0$. The linear fractional transformation is also known as a Moebius transformation [16].

The transformation functions are applied onto the swarm pattern which is polygonal in shape. Since the vertices of the polygonal pattern lie on the circle circumscribing the pattern, a circle to line and a line to circle transformations of the complex plane are used. However, the transformation function cannot be applied directly to the multi-robot pattern. This is due to the fact that the multi-robot pattern is defined on a global frame of reference while the mathematical function is applicable on the local frame of reference. Hence, the sequence of operations performed on the multi-robot pattern is:

*1) Transformation from global to local frame of reference*: The frame of reference of the multi-robot is temporarily transformed from the global to a local frame. The local frame of reference considered is such that the circumscribing circle is divided into four equal quadrants. Hence the centroid of the pattern lies on the origin position of the local frame.

*2) Discrete Transformation*: This step applies the mathematical transformation function on the multi-robot model. The transformation of a circle to a line is obtained from $w = i\frac{(1-z)}{(1+z)}$. Applying the equation on the Euclidean plane, the mapping function is deduced as $\left(\frac{2y}{1+x^2+y^2}, \frac{1-x^2-y^2}{1+x^2+y^2}\right)$. The transformation from a line to a circle is applied by considering a special case of the Moebius transformation. The transformation $w = \frac{1}{z}$ maps every straight line or circle onto a circle or straight line. It is also known as the inversion in the unit circle or reciprocal transformation. Applying the equation on the Euclidean plane, the mapping function is otherwise written as $\left(\frac{x}{x^2+y^2}, \frac{y}{x^2+y^2}\right)$. The destination coordinates obtained by the mathematical functions are the coordinates to which individual robot agents need to reposition while the pattern transforms. However, it is evident that these transformation functions are discrete in nature yielding only one set of destination coordinate rather than sub-goals or intermediate destination coordinates.

*3) Transformation from local to global frame of reference with magnification:* The destination coordinates are obtained on the local frame of reference. Hence, the local frame needs to be shifted to the global frame of reference. Since the mathematical functions considered in step (ii) are reducing functions (destination coordinates reduce the span of the pattern), a magnification ratio is used in the local frame to achieve gain in the destination coordinates.

*4) Path planning by discretization:* Since the achieved destination coordinate set is discrete, the major challenge in repositioning agents is to plan their path to the destination coordinates. In this paper, the technique adopted to reposition robots is along straight line trajectories without collisions. The straight line path between the agent and its estimated destination is discretized. A straight line discretization process is done by slicing the domain values to extrapolate the range values. This relates to the underlying principle of Discrete Event Simulations (DEVS). The potential of DEVS in path planning for robots is reported in [20].

## VI. SIMULATION STUDIES

Simulation studies were developed to validate and visualize the proposed geometric approaches for pattern formation and transformation. Most robotic simulators proved ineffective for incorporating the geometric approach. Hence, a non-robotic particle physics simulation engine was employed. The remainder of this section is organised into understanding the experimental environment, studies on pattern formation and transformation.

### A. Experimental Environment

The feasibility of the proposed approach was validated on the Processing [21] and Traer Physics [22] environment. Processing is an open source programming language and environment enabling visualizations for learning and prototyping. Traer Physics is a particle physics simulation engine for Processing.

The traer physics library has provisions for modeling a particle system, particles, springs and attractive or repulsive forces [22]. The particle system enables prototyping particles and forces. Particles represent objects having four properties, namely mass, position, velocity and age. Particles can be stationary or dynamic in an environment. Springs can connect two particles and prevent collisions. Springs are characterized by three properties, namely rest length, strength and damping. Attractions or repulsions pull particles together or apart and have two properties, namely strength and minimum distance. The simulations reported in this paper employ particle system, particles and attractive or repulsive forces.

The swarm pattern is designed as particles in an open environment with forces, namely macro and micro level forces of attraction and repulsion acting on the pattern. The macro level forces include repulsive forces, which act on the centroid of the swarm. The forces of repulsion are generated from obstacles (modeled as forces) in the environment. All robotic agents align themselves around the centroid with respect to the forces forming a virtual structure polygonal pattern. Obstacles in the path of the pattern are detected by the computation of the net force acting on the group of robots. Beyond a maximum threshold value of force, the pattern reacts appropriately by transforming its shape to avoid obstacles. The pattern regains its polygonal shape when the net force acting on the centroid decreases below a minimum threshold value, such as when the pattern has escaped from obstacles. The inter-agent bonding force and the forces of interaction with the centroid contribute to the micro level forces. The pattern generates a propulsive force to trace paths against repulsive forces.

The experimental setup comprised a tunnel through which the swarm had to displace. The walls of the tunnel generated repulsive forces and acted as the obstacle. The swarm initiated its motion from the left of the tunnel and aimed to reach a goal beyond the tunnel on the right side.

### B. Studies on Pattern Formation

Regular pattern formations were studied when the robot pattern displaced through obstacles similar to bridges and tunnels. The pattern was expected to deflate when subjected to a potential above a minimum threshold force, continue motion and inflate beyond the obstacles. Simulation results for regular pattern formation with 5 and 6 robots are presented in figure 2 (left columns).

Irregular pattern formations were studied when the geometric robot patterns displaced through obstacles that converged and hence offered a narrowed path of movement. This replicated motion through a funnelled path. The pattern was expected to deflate laterally or longitudinally when subjected to a potential above a minimum threshold. Simulation results for irregular pattern formation with 5 and 6 robots are presented in figure 2 (right columns).

It is observed from both the regular and irregular pattern formation simulations that the patterns deflated to traverse through obstacle paths. Beyond obstacle paths, the patterns inflated to achieve their original configuration. It is notable that obstacle avoidance is an inherent property of the system and hence implicitly guaranteed since obstacles are modelled as forces. These observations are consistent with the theoretical studies of Section 3 and according to the authors expectations. Hence, the simulation studies confirm the formulations of the geometric approach for polygonal robot configurations.

### C. Studies on Pattern Transformation

Both transformation methods discussed in Section IV and V were implemented. The macroscopic method of Section IV, consisting of a sequence of three operations, was implemented. Firstly, the swarm model was rotated to avoid collisions while deflating. Table I illustrates the different rotation angles that were applied on the swarm. Higher value angles resulted in collisions for most patterns. Angles less than 15 degrees proved effective for collision avoidance. Secondly, the macroscopic parameters were varied. This variation resulted in deflation or inflation of the pattern (along the x or y axis). Thirdly, a corrective rotation was applied to avoid agents from colliding against each other. Hence by transformation of the pattern, the swarm successfully displaced itself through the obstacle path. Figure 3 is a snapshot of the simulation studies for n = 3 to 6, 10 and 20 robots transforming in accordance with the first method.

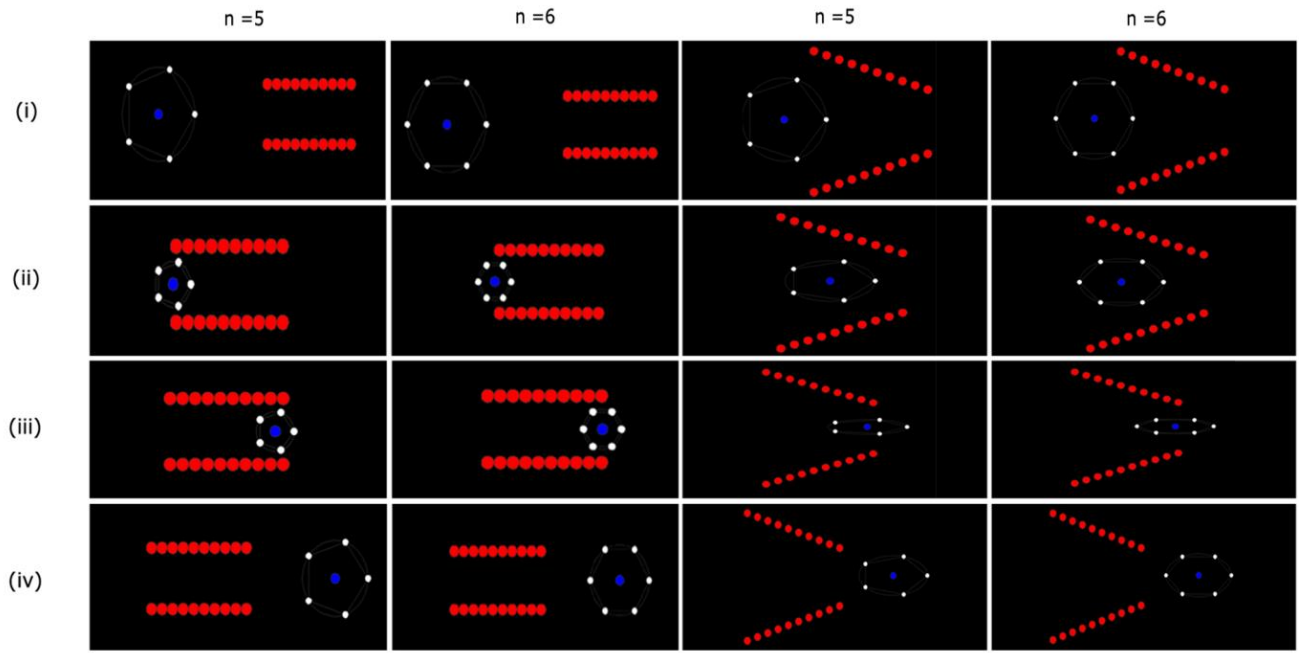

Fig. 2. Left columns (n = 5 and 6) Regular pattern formation and obstacle avoidance through a tunnel. Right columns (n = 5 and 6) Irregular pattern formation and obstacle avoidance through a funneled path.

TABLE I
PRE DEFINED ROTATION VALUES & ESTIMATED COLLISIONS

| Angle Offset / No. Of Robots | 15° | 30° | 45° | 60° |
|---|---|---|---|---|
| 3 | - | - | - | 1 |
| 4 | - | - | 2 | - |
| 5 | - | 1 | - | - |
| 6 | - | 3 | - | 3 |

The mathematical method of Section V which consists of a sequence of four steps was implemented. Firstly, the swarm pattern was transformed from the global to a local frame such that the centroid of the pattern lies on the origin of local frame of reference. Hence, the pattern is equally spanned over the four quadrants in the local frame of reference, which was necessary for proper implementation of the transformation functions.

Secondly, the discrete transformation function was applied on the microscopic property, namely the position coordinates of the individual robots in the pattern. The transformation from a circle to line was employed in order for the pattern to pass through the tunnel in the environment. Beyond the obstacles, the transformation from a line to circle was employed. Both transformation operations yield a set of discrete destination coordinates for each robot.

Thirdly, transformation from the local to global frame of reference was performed. The destination coordinates obtained in the local frame of reference were such that the pattern radius is reduced. Hence a magnification of the coordinates in the local frame was performed and further mapped on to the global frame of reference.

Fourthly, path planning by discretization was executed. This step is essential to determine the sub goals or intermediate position coordinates. Repositioning the robots to sub-goals or intermediate coordinates is a computationally expensive process. Straight line trajectories from agents to calculated destination coordinates without collisions were considered in the work reported in this paper. Figure 4 is a snapshot of the simulation studies for 17 robots that transform shape in accordance with the second method.

It is observed that the circle to line transformation yielded a pattern in which robotic agents were loosely equidistant. The line to circle transformation employing the reciprocal transformation yielded a polygon irregular in nature. This was due to the nature of the reciprocal transformation, which was anticipated.

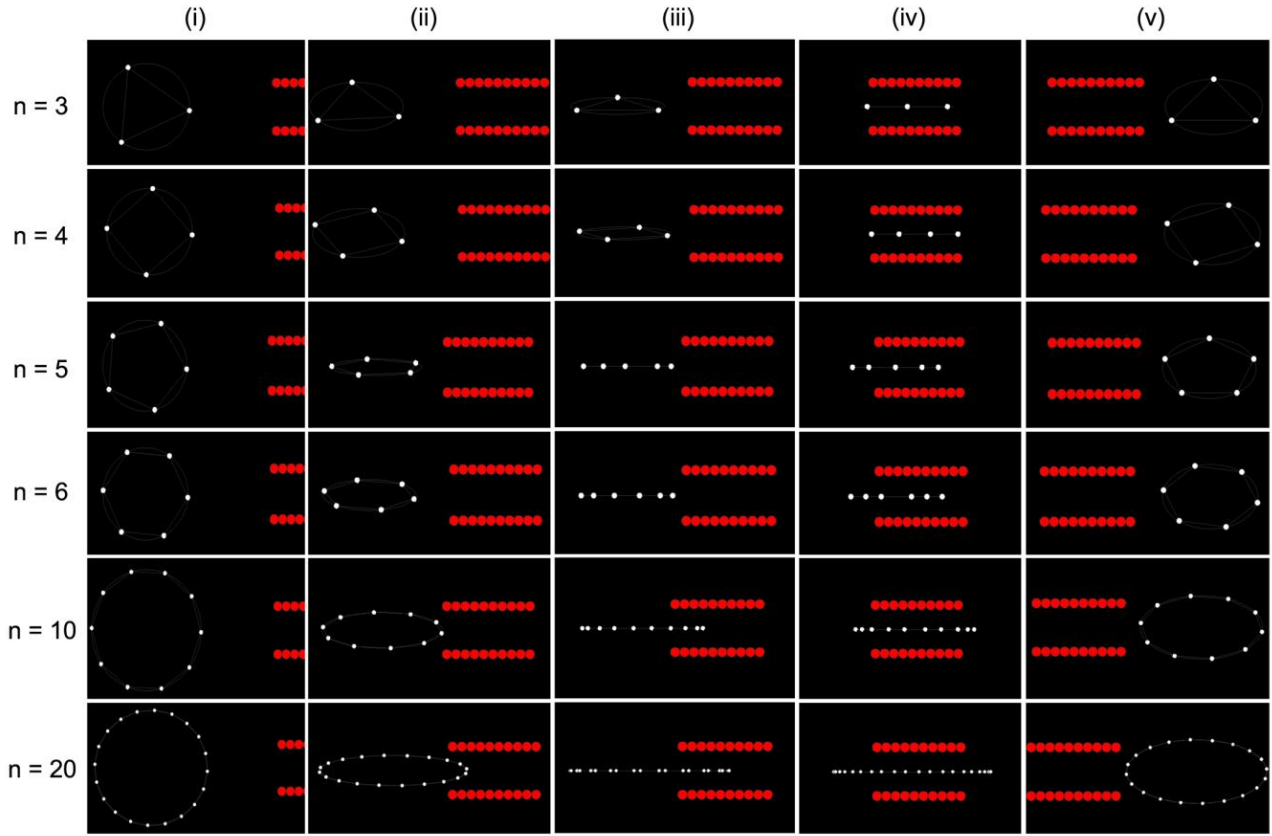

Fig. 3. Simulation results on Traer Physics and Processing simulator for the first method. (i) Rotated swarm model for various number of robots, (ii) Deflation of the model along the y – axis (For n = 10 and 20, inflation along x – axis performed), (iii) Transformed pattern without corrective rotation measure, (iv) Transformed pattern after corrective rotation measure is applied (Except for n = 3 and 4, since equidistance is more or less achieved), (v) Inverse transformation by inflation back to original pattern.

It was observed that in both methods, the swarm successfully displaced itself through the obstacle path by transforming shape. The transformed patterns were loosely equidistant. Collision avoidance between repositioning agents is not implicitly guaranteed. Hence, at least one operation in both methods ensured collision avoidance. The geometrical transformation of a circle to a line in both cases was achieved by transforming a regular polygonal pattern to an irregular pattern by repositioning agents. The observations are consistent with the theoretical studies in Section IV and V and according to the authors expectation.

### D. Results

The time taken to transform a pattern in the two transformation methods for different number of robots in the configuration was measured. This experiment was carried out for different number of robots varying from 3 to 25 and keeping the initial formation radius of the swarm a constant. Figure 5 (left) and Figure 5 (right) are graphs plotted using MATLAB and are based on the results obtained from simulation for the first and second transformation method respectively. The graphs show the time taken for transformation versus the number of robots in the pattern.

The average time taken to transform a pattern in the first transformation method was computed as 20.35 seconds. It is noted that the values plotted on the graph can be divided into two bands. Firstly, the set of times that lie below 20 seconds and secondly, those that lie above 20 seconds. The first band follows a linear trend with a steady rise. Though the second set of times is scattered, they form three different clusters for n = 4 to 8, 11 to 15 and 22 to 25.

Working from left to right on the robot axis, the smallest time taken for transforming the pattern was observed when three robots constituted the pattern. This was due to the fact that a corrective rotation step was not required for obtaining equidistance between the patterns. Hence, the robots traversed lesser distances to achieve the transformed pattern. There is a steep rise in the time taken to transform the patterns for n = 4 to 8. This can be accounted for by the fact that the initial rotation and further rotation steps cause the robots to trace further distances to keep themselves equidistant.

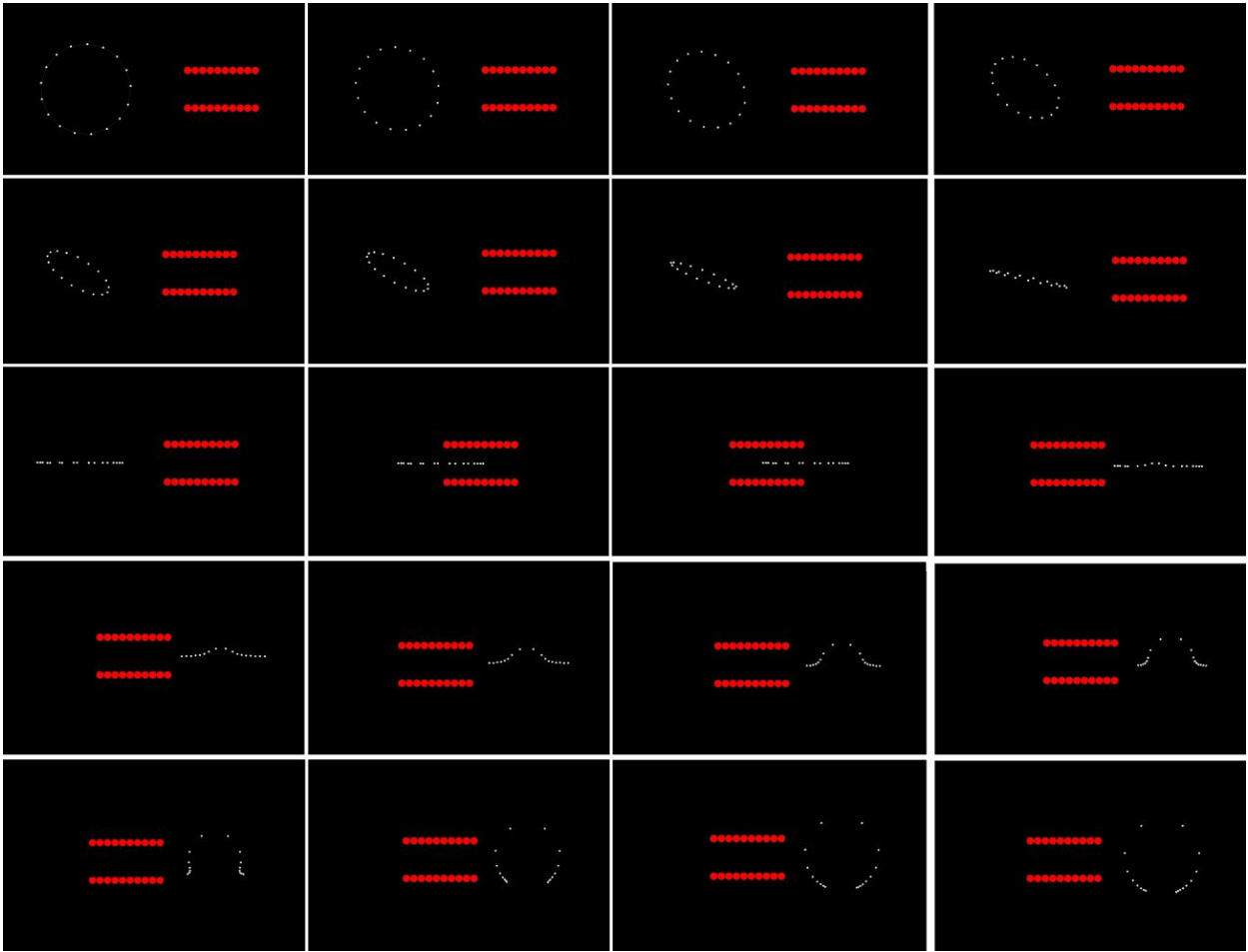

Fig. 4. Simulation studies on Processing and Traer Physics for the second transformation method presented in two stages. . Firstly, a circle to a line transformation (First three rows excluding the fourth sequence in the third row). Secondly, a line to circle transformation (Fourth sequence of the third row and last two rows).

For n = 9 and 10, however the time taken has a steep fall. It is likely that the robots trace less distances for collision avoidance in these cases. The highest time for transformation was noted for n = 11 to 15. This was due to the fact that a higher degree of rotation was required initially to avoid collisions. Hence the robots took longer times to reposition. In the further rotation step, more agents compared to the previous cases had to reposition. Hence, this accounts for this cluster having the largest time. For n = 16 to 21 there is a gradual increase in the time taken to transform, but much lower than the time taken by the previous cluster since the corrective rotation step was not necessary. An increase in the time taken for n = 22 to 25 is noted. This was due to the fact that during transformation an inflation operation was performed along the x-axis to accommodate all the robots. Hence, this led to an increase in time.

The average time taken to transform a pattern in the second method was computed as 21.40 seconds, slightly higher than the first method. It is noted that the values plotted on the graph can be divided into two bands. Firstly, the set of times that lie above 21 seconds and secondly, those that lie below 21 seconds. The first band follows a linear trend with a steady decrease. Though the second band lies scattered, all set of times in this band lie close to 20 seconds.

Working from left to right on the robot axis, the longest time taken to transform a pattern was observed when 3 robots constituted the pattern. This was due to the nature of the mathematical transformation in which the 3 robots traverse longer distances within the pattern to reach their destination coordinates. The transformation time decreases steadily until n = 11 due to the fact that the distance traversed within the pattern decreases. For n > 11, the set of times is scattered closely around the 20 second time line. It is evident that the mathematical transformation method is consistent and effective for patterns with more than 11 constituting robots. This is due to the fact that robots traversed lesser distances in the pattern. However, the time taken for a few configurations (n > 11) is greater than 20 seconds. This can be accounted for the fact that collision avoidance in robots was achieved by temporarily decelerating alternative robots in the configuration and hence led to a small increase in time. The lowest time taken to transform is for n = 20. In this case, the distance traversed by robots and collision avoidance deceleration of alternative robots in the pattern is minimal.

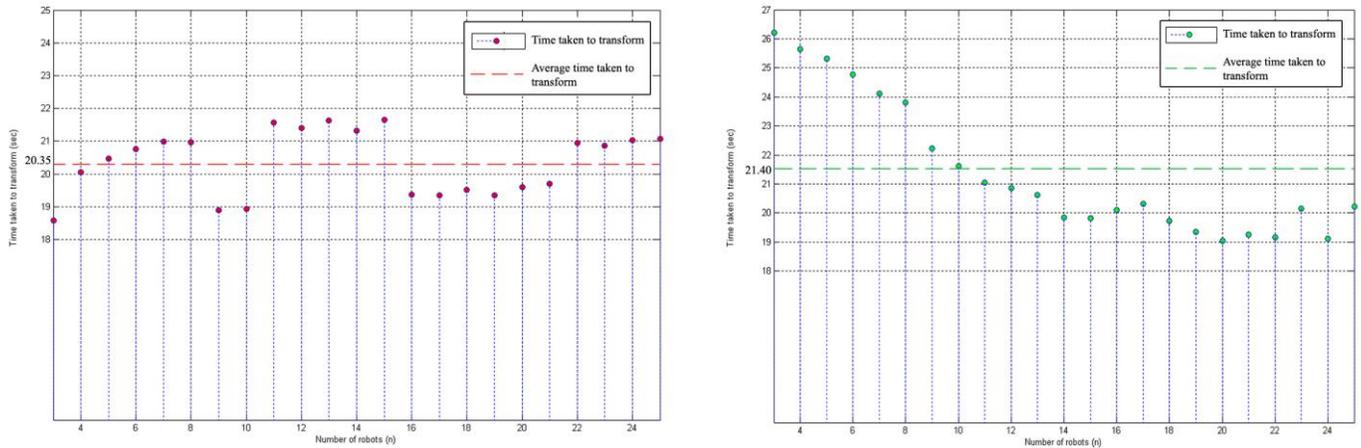

Fig. 5. Graphs obtained from experiments. (Left) Graph plotted based on the first transformation method. (Right) Graph plotted based on the second transformation method.

It is understood from the graphs that the mathematical transformation method employing both macroscopic and microscopic parameters is not advantageous for small number of robots. For smaller number of robots, the robots in the mathematical method traverse more distance within the pattern. As the number of robots increase, the distance traversed by a robot within the pattern decreases. The mathematical method performs better than the macroscopic method for higher number of robots. As the number of robots increase, the mathematical method tends to be effective since the time taken to transform decreases. However, the first method performs consistently for any number of robots.

### E. Comparing the methods

The transformation methods presented in this paper are feasible methods for reconfiguring patterns. However, it is noted that the mathematical method employing Moebius transformation is not strictly macroscopic in nature. The microscopic properties of the swarm units are taken into consideration. For example, path planning of individual robots is necessary to reposition the robots. The method is not advantageous for small number of robots in the pattern. Moreover the mathematical transformation method is a discrete transformation method. Hence, discretizing and quantizing the path to reposition are required. This is a computationally expensive process unsupported and unwarranted on minimal processing swarm units. Therefore global planning is required thereby increasing wireless communication overheads. A high bandwidth for communication and synchronized and consistent communication with a centralized unit are challenges in realizing the mathematical method in real time.

On the other hand, the macroscopic method considers the group behaviour of the swarm system. Hence, individual robots need not be addressed, eliminating microscopic parameter operations. For example, transformation in the first method is obtained by a sequence of operations performed on the entire swarm pattern rather than considering individual robot path planning. The macroscopic method is observed to be consistent in the time taken for transformation, and is also a continuous method thereby reducing computations for individual robot planning. This method would hence offer better synchronization between the swarm units since local planning is sufficient. Hence wireless communication overheads are relatively less compared to the mathematical method.

By implementing a macroscopic method in a real time robot system, planning overheads for individual robots could be minimized. However, a mathematical transformation function is advantageous since it belongs to an analytical class of tools and mathematical analysis is possible.

In summary, the simulation studies confirm the feasibility of the proposed methods. The transformation cases discussed in Section III are considered in the transformation methods. A brief comparison between the method employing only macroscopic parameters and the method employing both microscopic and macroscopic parameters is presented based on the results obtained.

### VII. CONCLUSIONS

The swarm pattern formation model presented in this paper considers macroscopic and microscopic primitives. The swarm pattern transformation model comprising two transformation methods, namely a macroscopic transformation and mathematical transformation method are proposed. The first method is a macroscopic parameter method while the second considers both, macroscopic and microscopic parameters. A formal definition for transformation is presented with four special cases of transformation. Elementary and geometrical transformations are considered by repositioning agents. Transformation using both methods is achieved by a sequence of operations performed on the swarm pattern. The proposed methods are implemented on the Processing and Traer Physics environment. A comparison between the two methods considering transformation time from one pattern to another is presented. The simulation studies confirm the feasibility of the proposed methods.

Future work will include the real time implementation of the proposed transformation methods on a swarm robot system. The challenges in mapping simulation studies to real time robot systems will be studied. Efforts will be made to explore continuous mathematical transformation methods which are expected to minimize individual robot path planning.